# Training Humans and Machines

**Aki Nikolaidis PhD** (aki.nikolaidis@childmind.org)
The Child Mind Institute, 445 Park Avenue. New York, NY 10022

**Abstract:**

**For many years, researchers in psychology, education, statistics, and machine learning have been developing practical methods to improve learning speed, retention, and generalizability, and this work has been successful. Many of these methods are rooted in common underlying principles that seem to drive learning and overlearning in both humans and machines. I present a review of a small part of this work to point to potentially novel applications in both machine and human learning that may be worth exploring.**

## Introduction

Researchers have long sought to bridge the divide between human and machine intelligence, developing neural network[1], Bayesian[2], and other models[3] of human cognition. In practice, training humans and machines is fraught with many difficulties that have proven difficult to overcome. For many years, researchers and practitioners in both the human and machine learning fields have endeavored to develop new methods of improving learning quality and its generalization. While these two communities have made considerable progress in creating experimental methods to improve learner performance, they have identified and tackled similar problems without benefitting enough from the input of the other. We examine some ways in which learning problems for machines and humans are similar, and practical methods for improving learner performance in both humans and machines that may be worth exploring across disciplines.

## Building Learning Models

The stages of skill learning in humans have similar correlates in machine learning. Skill acquisition is split into three phases: First, the Cognitive phase, during which skill performance is slow, highly variable, and requires considerable executive function to support task acquisition[4]. Here, patterns of corticostriatal activity that form the stimulus-response unitization of the task[5], and this acquisition is regulated by executive networks such as the salience and cingulo-opercular networks[6]. Next, in the Associative stage, performance is more reliable, yet it is also fluid and less executive function is required. Corticostriatal function is still essential for execution of the skill during this stage, though as some parts of the task become offloaded to cortex and are automated, cortical processing becomes more important. Finally, in the Autonomous stage, performance is accurate and efficient, and little to no cognitive involvement is required. The task becomes highly specialized and inflexible, and slight perturbations in task-presentation will lead to drops in performance. In this stage, the corticostriatal system is no longer involved, and the skill has been automated and completely deposited in cortex[7]. This is exemplified by studies demonstrating that lesions to the striatum in rats selectively prevents the acquisition and accurate performance of any stimulus-response pairings in the Cognitive or Associative stages, but does not impact Autonomous, or overlearned, behaviors[8]. Some of our work has also implicated these corticostriatal circuits in complex skill acquisition and generalization in humans as well[9,10].

## Overlearning and Overfitting

Broadly, overfitting is a process by which a model becomes fit to a training set to the point where the model is complex enough to fit noise in the training set, resulting in worse performance in an out-of-sample set of data[11]. Overlearning, on the other hand, is the process by which a learner develops a skill to the point that the skill becomes automated, and very minor perturbations in the task can lead to significant decreases in performance[7]. These two domains are heavily studied aspects of human and machine learning. Understanding the extent to which the learning process from underfitting to overfitting is mirrored by these three phases of human skill acquisition is an important step for tying together human and machine learning research, and there are substantial parallels. For example, during the Cognitive phase, an underfit model of low complexity is constructed by the corticostriatal system, leading to lower accuracy and high variability in performance. As training continues and the skill becomes further unitized by the corticostriatal system, the learner reaches a peak of performance accuracy in both training and test sets. This stage represents a maximization of task flexibility that is similarly seen during the Associative stage of skill acquisition. Finally, during the Autonomous stage, as in the case of overfitting, the learner's model has grown so large it fits the noise in the training data to the extent that the model becomes inflexible to new data, and test-set accuracy drops. The patterns of under-overfitting seem, at least at first glance, to have substantial overlap in human and machine learners. This, is most clearly demonstrated in independent fields of research showing that introducing variability into the training process has a beneficial impact on learning models in both humans and machines.

## Training for Variability

### In Humans

Studies have demonstrated that introducing variability in the learning process can impact many aspects of the learning process. In the study of skill acquisition, researchers have demonstrated that changing the attentional focus of the trainee and its impact on skill acquisition. They showed changing the focus of the trainee from the task as a whole (Fixed Priority- FP) to subparts of the task in the context of performing the task as a whole (Variable Priority- VP) has a dramatic impact on the acquisition of the task[12]. In one training study, participants were given either the VP or FP training strategy in the acquisition of a complex cognitive and motor videogame, Space Fortress. FP and VP training was equally beneficial for participants with high initial performance, showing the same levels of performance improvement over twenty hours of training. With poor initial performers however, the VP training was so successful that after training they reached the same level of performance as the high performers had reached over 20 hours of training. On the other hand, the FP training was particularly difficult for the poor performing group, showing minimal skill improvement over the course of training[12].

Another study demonstrated that this training technique is capable of overcoming age-related performance deficits as well. In a complex motor task where older adults tend to perform significantly poorer compared to young adults, groups of both young and old adults were trained with either a FP or VP training strategy. By the end of training, the older adults training with the VP were not only as good as the young adults by the beginning of training, but they matched the performance of the young adults after training as well[13]. This work suggests that in the learning of highly complex skills, learners for whom the task is quite difficult, benefit from tackling subparts of the task, and can be overwhelmed by the task as a whole at first. Furthermore, VP training has also been demonstrated to create more generalizable representations of the trained skills and improve transfer in dual-task training[14]. Computational models of the basal ganglia and cortico-striatal interactions in skill learning have also provided a theoretical framework for how FP training leads to overlearned and less flexible skill representations compared to VP training[15].

Varying the training and feedback schedule also the impact have on learning and transfer. For example, training with randomized blocks of stimuli, leads to slower performance over the course of training as compared to fixed blocks of stimuli[16]. Randomized training also led to better performance after a 10-day retention test in both the fixed and randomized retest conditions, whereas the fixed block training led to much worse performance in the randomized block retest condition, indicating fixed training had led to development of overlearned and inflexible skills. This effect has been widely replicated in real world skills including keyboard skills[17], and badminton[18].

The same pattern of results have been found in training experiments where the feedback schedule varied. In one study, learners received performance feedback after every 1, 5, or 15 trials[19]. Training error curves descended most quickly for trainees in the 1-trial feedback condition, more slowly for the 5-trial, and even more slowly for the 15 trial feedback conditions. By the end of six training acquisition blocks, all three conditions showed approximately equivalent performance, and after a 10-minute retention test, performance had decayed equally for all conditions. However, after a 2-day retention test, the performance of was highest for the 15-trial feedback condition, and worst for the 1-trial feedback condition, demonstrating that blocked feedback conditions had a significant impact in creating more generalizable acquisitions of the skill[20]. Similar effects have been found in a range of other motor training[21], name learning[22], and computer programming[23] studies that manipulate feedback schedules. Generally, it seems slowing learning leads to greater generalization, and in human learning, a variety of methods have been developed to demonstrate this effect. Machine learning research, on the other hand, has developed alternative methods for increasing variability in the training process.

### In Machines

There are several methods for increasing variability in the training process towards the end of improving generalizability in machine learning models. Perhaps one of the most popular areas of research in this domain are ensemble methods. Ensemble learning refers to a set of methods in machine learning by which a large group of learners are all deployed and by aggregating the models learned by these base learners, ensemble methods are able to reach higher performance and generalizability in their models. This method effectively increases the scope of the training set that the learners are exposed to, thereby increasing the range of models that are possible to learn, and creating an aggregate model of all these base models. One popular form of ensemble learning involves bootstrap aggregation, or bagging[24]. In this method, models are applied across many bootstraps of the samples and then averaged together. Bagging enables the final model to experience a greater range of solutions than would be considered with only a single learner alone. Many bootstraps of the observations are created and models are created for each one and then aggregated to create an average model[25].

Similarly, boosting is a machine learning technique where a range of weaker models that perform only slightly better than chance are subjected to weighted aggregation, based on their error in the training set, to form a single superior learner. Boosting and bagging have been shown to result in better predictive performance in supervised learning and less overfit models[24,26].

Neural networks are well known for their tendency towards overfitting, but with enough training samples these methods produced well-generalized and high performing models. One popular method for enhancing performance in neural network research and practice is increasing the size of the training set. One popular method is creating altered, skewed, and noisy copies of the data to train on and mixing them into the training set to substantially increase both the size and variability of the training set. This has been shown to have significant impacts on prediction error and preventing overfitting[27]. Adding noise directly into the training set, called noise injection, is another popular method for creating variability in the training data, and is especially popular in neural networks. Recent work has demonstrated that adding noise to training data may reduce the tendency of neural networks to overfit to the training sample[28]. These methods of adding variability to the training data in neural networks seem to generalize to other learning methods as well, and have been implemented in other algorithms and in bioinformatics problems as well[29].

**Transfer in Humans and Machines**

Transfer of learning has been studied in the human psychology literature for over 100 years[30], and refers to the ability of what is learned in a training to be deployed in other contexts[31], such as different times, places, tasks, orders, or many other forms[32]. The most widely held theory in this domain is known as the theory of identical elements: for skill A to transfer to skill B, A and B must share some identical elements that enables the transfer of the skill production. Broadly speaking, training and testing a machine learning model is a form of transfer assessment as well, and in this literature, people discuss the test set as being representative of the training set in much the same way that the human literature discusses identical elements between skills. Much more recently, the machine learning literature has focused on transfer learning more specifically, with the creation of many methods designed specifically to provide generalizable models that are trained to handle many tasks in one. Transfer learning is often used in deep learning, where pre-trained models are used as starting points in computer vision or language processing tasks, given the massive computational resources required to develop robust neural network models for these domains. Inductive transfer is the concept that the search space of possible models has a smaller range of allowable hypotheses, allowing for better initial performance and faster improvements in performance during training. The problem of transfer remains a substantial obstacle in both human and machine learning. These fields both have an extensive array of non-overlapping methods for manipulating the training and learning experience. They may work together for mutual benefit of studying and deploying more advanced training paradigms and learning agents to enhance learning and transfer. For instance, the human learning research has extensive experience with adjusting the training process in many ways that could be modeled and applied successfully in the machine learning literature; on the other hand, machine learning research has substantial experience in creating different types of learners, and demonstrating their relative effectiveness given different learner tasks, which could be an important method for modeling natural variation in human learners.

**Conclusions & Future Directions**

Of course, human and machine learning have more differences than similarities, but teaching human and machines to perform a task comes with a range of related challenges. In some regards both humans and machine learners face the same challenges, building models that improve performance but remain flexible enough to prevent overlearning, transferring previously learned productions to new contexts when appropriate. Both humans and machines seem to learn more slowly when more variable data are presented during training, and the resulting models developed are more generalizable.

Moving forward, we make a few suggestions for each field. Broadly speaking for human learning, this field might benefit from strategies common in machine learning creating and manipulating training data and methods of learner exposure to the training data. For example, we suggest that human learning may benefit from developing novel methods for generating altered or noisy training data to encourage the development of generalizable skill productions. Furthermore, human learning may benefit from developing efforts to methodologically drive learners back and forth between stages of learning, as can be in machine learning by using more or less complex models.

Generally, machine learning may benefit from strategies common in human learning of manipulating the framework of the training to enhance learning and generalization. For instance, particularly difficult problems may find benefit from exploring how training strategy, or 'model focus', may be varied over the course of training to improve learning, as in the case of the FP and VP training strategies in humans. Furthermore, machine learning research may find benefit in developing methods of manipulating feedback schedules to reduce learner sensitivity to error in prediction. There are likely many more areas of overlap, and methods used in one domain that may be of benefit to the other. We suggest an expanded and continued conversation.

**References**

1. McClelland, J. L., McNaughton, B. L. & O'Reilly, R. C. Why there are complementary learning systems in the hippocampus and neocortex: insights from the successes and failures of connectionist models of learning and memory. *Psychol. Rev.* **102,** 419–457 (1995).
2. Griffiths, T. L., Kemp, C. & Tenenbaum, J. B. Bayesian models of cognition. (2008).
3. Waskan & Jonathan, A. *Models and Cognition*. (MIT Press, 2012).
4. Wulf, G. & Lewthwaite, R. Effortless Motor Learning?: An External Focus of Attention Enhances Movement Effectiveness and Efficiency. in *Effortless Attention* (The MIT Press, 2010).
5. Wulf, G. *Attention and Motor Skill Learning*. (Human Kinetics, 2007).
6. Petersen, S. E., van Mier, H., Fiez, J. A. & Raichle, M. E. The effects of practice on the functional anatomy of task performance. *Proceedings of the National Academy of Sciences* **95,** 853–860 (1998).
7. Doyon, J. *et al.* Contributions of the basal ganglia and functionally related brain structures to motor learning. *Behav. Brain Res.* **199,** 61–75 (2009).
8. Eichenbaum, H. & Cohen, N. J. *From Conditioning to Conscious Recollection: Memory Systems of the Brain*. (Oxford University Press, USA, 2004).
9. Nikolaidis, A., Voss, M. W., Lee, H., Vo, L. T. K. & Kramer, A. F. Parietal plasticity after training with a complex video game is associated with individual differences in improvements in an untrained working memory task. *Front. Hum. Neurosci.* **8,** (2014).
10. Nikolaidis, A., Goatz, D., Smaragdis, P. & Kramer, A. Predicting Skill-Based Task Performance and Learning with fMRI Motor and Subcortical Network Connectivity. *2015 International Workshop on Pattern Recognition in NeuroImaging* 93–96 (2015).
11. Babyak, M. a. What You See May Not Be What You Get: A Brief, Nontechnical Introduction to Overfitting in Regression-Type Models. *Psychosom. Med.* **66,** 411–421 (2004).
12. Voss, M. W. *et al.* Effects of training strategies implemented in a complex videogame on functional connectivity of attentional networks. *Neuroimage* (2011). doi:10.1016/j.neuroimage.2011.03.052
13. Kramer, A. F., Larish, J. L., Weber, T. A. & Bardell, L. Training for Executive Control: Task Coordination Strategies and Aging. (1999).
14. Kramer, A. F., Larish, J. F. & Strayer, D. L. Training for Attentional control in dual task settings: a comparison of young and old adults. *Journal of Experimental ...* (1995).
15. Fu, W.-T., Rong, P., Lee, H., Kramer, A. F. & Graybiel, A. M. A Computational Model of Complex Skill Learning in Varied-Priority Training. 1482–1487 (1995).
16. Shea, J. B. & Morgan, R. L. Contextual interference effects on the acquisition, retention, and transfer of a motor skill. *J. Exp. Psychol. Hum. Learn.* **5,** 179 (1979).
17. Baddeley, A. D. & Longman, D. J. A. The Influence of Length and Frequency of Training Session on the Rate of Learning to Type. *Ergonomics* **21,** 627–635 (1978).
18. Goode, S. & Magill, R. A. Contextual Interference Effects in Learning Three Badminton Serves. *Res. Q. Exerc. Sport* **57,** 308–314 (1986).
19. Schmidt, R. A., Young, D. E., Swinnen, S. & Shapiro, D. C. Summary knowledge of results for skill acquisition: support for the guidance hypothesis. *J. Exp. Psychol. Learn. Mem. Cogn.* **15,** 352–359 (1989).
20. Schmidt, B. R. A. & Bjork, R. A. NEW CONCEPTUALIZATIONS OF PRACTICE : Common Principles in Three Paradigms Suggest New Concepts for Training. 207–218 (1992).
21. Wulf, G. & Schmidt, R. A. The learning of generalized motor programs: Reducing the relative frequency of knowledge of results enhances memory. *J. Exp. Psychol. Learn. Mem. Cogn.* **15,** 748 (1989).
22. Landauer, T. K. & Bjork, R. A. Optimum rehearsal patterns and name learning. *Practical aspects of memory* 625–632 (1978).
23. Corbett, A. T. & Anderson, J. R. Feedback control and learning to program with the CMU LISP tutor. *Department of Psychology* 28 (1991).
24. Breiman, L. Bagging predictors. *Mach. Learn.* **24,** 123–140 (1996).
25. Zhou, Z.-H. *Ensemble Methods: Foundations and Algorithms*. (CRC Press, 2012).
26. Freund, Y., Schapire, R. E. & Others. Experiments with a new boosting algorithm. in *Icml* **96,** 148–156 (Bari, Italy, 1996).
27. Simard, P. Y., Steinkraus, D., Platt, J. C. & Others. Best practices for convolutional neural networks applied to visual document analysis. in *ICDAR* **3,** 958–962 (cs.cmu.edu, 2003).
28. Sietsma, J. & Dow, R. J. F. Creating artificial neural networks that generalize. *Neural Netw.* **4,** 67–79 (1991).
29. Yip, K. Y. & Gerstein, M. Training set expansion: an approach to improving the reconstruction of biological networks from limited and uneven reliable interactions. *Bioinformatics* **25,** 243–250 (2009).
30. Woodworth, R. S. & Thorndike, E. L. The influence of improvement in one mental function upon the efficiency of other functions.(I). *Psychol. Rev.* 247–261 (1901).
31. Burke, L. a. & Hutchins, H. M. Training Transfer: An Integrative Literature Review. *Human Resource Development Review* **6,** 263–296 (2007).
32. Barnett, S. M. & Ceci, S. J. When and where do we apply what we learn?: A taxonomy for far transfer. *Psychol. Bull.* **128,** 612–637 (2002).